\documentclass{IEEEtran}
\usepackage{graphicx}
\usepackage{balance}
\usepackage{xcolor}
\usepackage{cite}
\usepackage{array}
\usepackage{mathtools}
\usepackage{epsfig}
\usepackage{verbatim}
\usepackage{url}
\usepackage{cite}
\usepackage{graphicx}
\usepackage{caption}
\usepackage{subcaption}
\usepackage{amsmath}
\usepackage {lscape}

\begin{document}

\title{A Novel Falling-Ball Algorithm for Image Segmentation}

\author{Asra Aslam, Ekram Khan, Mohammad Samar Ansari, and M.M. Sufyan Beg 
	\thanks{A. Aslam is with the Insight Centre of Data Analytics, National University of Ireland, Galway, Ireland. e-mail: (asra.aslam@insight-centre.org).}
	\thanks{E. Khan, M.S. Ansari are with the Department of Electronics Engineering, Aligarh Muslim University, Aligarh, 202002 India. e-mail: (ekhan67@gmail.com, samar.ansari@zhcet.ac.in).}
	\thanks{M.S. Beg is with the Department of Computer Engineering, Aligarh Muslim University, Aligarh, 202002 India. e-mail: (mmsbeg@hotmail.com).}
}

\IEEEpubid{\begin{minipage}{\textwidth}\ \\ \\ \\ \\ [12pt]
\textit{Preprint submitted to Image and Vision Computing}.
\end{minipage}}

\markboth{}%
{}

\maketitle

\begin{abstract}

Image segmentation refers to the separation of objects from the background, and has been one of the most challenging aspects of digital image processing. Practically it is impossible to design a segmentation algorithm which has 100\% accuracy, and therefore numerous segmentation techniques have been proposed in the literature, each with certain limitations. In this paper, a novel Falling-Ball algorithm is presented, which is a region-based segmentation algorithm, and an alternative to watershed transform (based on waterfall model). The proposed algorithm detects the catchment basins by assuming that a ball falling from hilly terrains will stop in a catchment basin. Once catchment basins are identified, the association of each pixel with one of the catchment basin is obtained using multi-criteria fuzzy logic. Edges are constructed by dividing the image into different catchment basins with the help of a membership function. Finally, a closed contour algorithm is applied to find closed regions, and objects within closed regions are segmented using intensity information. The performance of the proposed algorithm is evaluated both objectively as well as subjectively. Simulation results show that the proposed algorithm gives superior performance over both the conventional sobel edge detection method and the watershed segmentation algorithm. 

\end{abstract}


\section{Introduction}
\label{sec:introduction}

With almost the entirety of image capturing and processing moving to the digital domain, the fields of digital image processing and machine vision have progressed tremendously over the past decade. Amongst the various digital processing operations that are frequently employed in such applications is `image segmentation,' wherein the objective is to alter and/or simplify the representation of a digital image into a more meaningful one, and consequently make it easier to analyze and process. Most commonly, image segmentation has been used to locate boundaries (curves, or edges/lines) in images, which leads to the segregation of objects and persons contained in the image. Typical applications of image segmentation can be seen in video surveillance systems \cite{jha2021real,snidaro2005video}, traffic monitoring \cite{pramanik2021real}, recognition tasks (face, iris, fingerprints), object detection \cite{zaidi2021survey,aslam2020object}, content-based image retrieval systems, sports analysis \cite{koundinya2012survey}, satellite imagery \cite{pal2002multispectral,bins1996satellite}, and medical imaging \cite{olabarriaga2001interaction,prajapati2015brain,murugavalli2007improved,aslam2015improved}. Image segmentation has proved to be a worthy tool towards locating tumors, estimating volumes of fluids/tissues, and surgery simulation.

\subsection{Motivation}

As mentioned above, segmentation is a procedure that divides an image into its constituent parts or objects. The level to which the division is carried depends on the specific problem. In practice, it is impossible to design an image segmentation algorithm which is capable of 100\% accuracy, and therefore the design of a high-performance segmentation algorithm is still an open problem. The technical literature contains many segmentation techniques, each with its own set of advantages and limitations. The main motivation behind this work is to present a novel image segmentation algorithm which outperforms the existing algorithms for the same task.

\subsection{Literature Review}

Many image segmentation algorithms have been developed in the recent past, with varied advantages and limitations being inherently associated with each of the proposals.
These are based on  the threshold approach \cite{prajapati2015brain,ahmad1999local}, region-growing approach like watershed \cite{dhage2015watershed,hojjatoleslami1998region,gambotto1993new,adams1994seeded}, fuzzy approach \cite{kerre2000fuzzy,chi1996fuzzy,tizhoosh1997fuzzy}, and graph-based methods \cite{wei2015medical} etc.

Some of these techniques are developed for a particular application whereas others work for all applications with lower accuracy. Most of the threshold-based segmentation approaches employ edge detection. Edges are calculated using intensity gradient or Laplacian operations. Some well-known edge detection operations are Sobel, Prewitt \& Roberts \cite{gonzalez2009digital}. Generally image gradients are compared with a threshold for edge detection. Selection of the threshold is an important factor, and significantly affects the performance of edge detection technique. Also the edges detected by these methods consist of thick lines. Hence, their is a requirement of reducing them to thin lines (single pixel width) by thinning. Suitable post-processing is therefore required to get the final edge pixels. Moreover, such methods are highly sensitive to noise.

The Marr-Hildreth edge detector \cite{gonzalez2009digital} uses a differential operator that can be tuned to any desired scale. It uses the Laplacian-of-a-Gaussian (LoG) operator for edge detection. A major disadvantage of this method is that the  LoG operator malfunctions at the corner and curves. Canny edge detection approach provides the following advantages: low error rate, well-localized edge points, and single-edge point response. However, these advantages come at the expense of increased algorithmic complexity.

One of the most successful algorithms in the region-growing class is the watershed segmentation algorithm. The concept of watersheds is based on topographic view of image, wherein the height of a mountain is proportional to the pixel intensity values. Thus, the points of higher intensity lie on the top of hills while the points of lower intensities form the valleys. Plateaus are formed at constant intensity areas.  Thereafter, the entire topography is flooded with water. As  the water level rises, different catchment basins will begin to merge. Prior to the merging of basins, dams are constructed at those points through dilation. Finally, these dam lines yield the edges. One of the most significant advantages of watershed algorithm is that it always results in closed contours. However, for complex images (like that of brain tumor,  or satellite images) it is over-sensitive and results in too many closed contours.

The application of Fuzzy Logic in digital image processing has  become an emerging branch of research due to the many benefits it offers in solving typical image-related problems. Fuzzy image processing has three main stages: image fuzzification, modification of membership values, and image defuzzification (optional). The coding of image data is referred to as fuzzification and decoding of results as defuzzification. These techniques convert image data into useful form for further processing of image with fuzzy techniques. Thereafter, suitable modification of membership values is done using one of the several available techniques like fuzzy rule-based approach, fuzzy clustering, fuzzy integration approach etc.

\subsection{Contribution}

Existing segmentation algorithms have limitations such as  dependency on threshold selection, dependencies on thinning algorithms, inability to form closed contours, and non-optimal sensitivity levels. These limitations point to the fact that there is still much research work required in this area. The proposed algorithm is an attempt in the direction of removing these limitations. 

Although the proposed approach also works on the principle of watershed catchment basins (as does the watershed segmentation algorithm), the most significant difference is that the topology is not flooded with water from the bottom as is done in the watershed algorithm. When the topology is filled from the bottom, a water droplet reaching the peak points of the topology can divide into multiple smaller droplets and these droplets may end up in different catchment basins. The proposed algorithm fills the topology with water but the water is now filled from the top (like rain). Each drop of water is modelled as a falling ball, and since a ball cannot break into multiple smaller portions, each drop (ball) can only end up in one catchment basin. Therefore, the behavior of the falling ball is probabilistic in nature. Thereafter, the  proposed approach employs fuzzy logic to make the decision for the falling ball to go into one of the many optional catchment basins. Further, a closed-contour algorithm is applied to find closed regions in the image. Finally, objects can be extracted from the image. Experiments are performed on a set of images. Comparative analysis is performed, by comparing the performance of the proposed algorithm with existing algorithms in terms of gray level uniformity measure (GU), Q-parameter \& relative ultimate measurement accuracy (RUMA). Results of these comparative analyses demonstrate  the superiority of the proposed method.

The remainder of the paper is organized as follows. A brief overview of existing segmentation techniques is presented in Section--\ref{sec:background}. Details of the proposed Falling-Ball algorithm are presented in Section--\ref{sec:proposed}. Algorithmic implementation is discussed in Section--\ref{sec:implementation}. Simulation results and comparisons are given in Section--\ref{sec:comparison}. Lastly, concluding remarks and avenues for future work appear in Section--\ref{sec:conclusion}.


\section{Background}
\label{sec:background}

Image segmentation algorithms are typically based on one of two basic properties of intensity values: discontinuity and similarity.
\begin{itemize}
	\item In the first category, the approach is to partition an image based on abrupt changes in intensity. Edge-based segmentation  is the principal approach in this category \cite{gonzalez2009digital}.
	\item The approaches in the second category are based on partitioning an image into regions that are similar according to a set of predefined criteria.  Region-based segmentation  is the main approach in this category \cite{gonzalez2009digital}.
\end{itemize}

\subsection{Edge Detection based Segmentation}

\begin{figure}[hb]
\begin{center}
\includegraphics[width=0.4\textwidth]{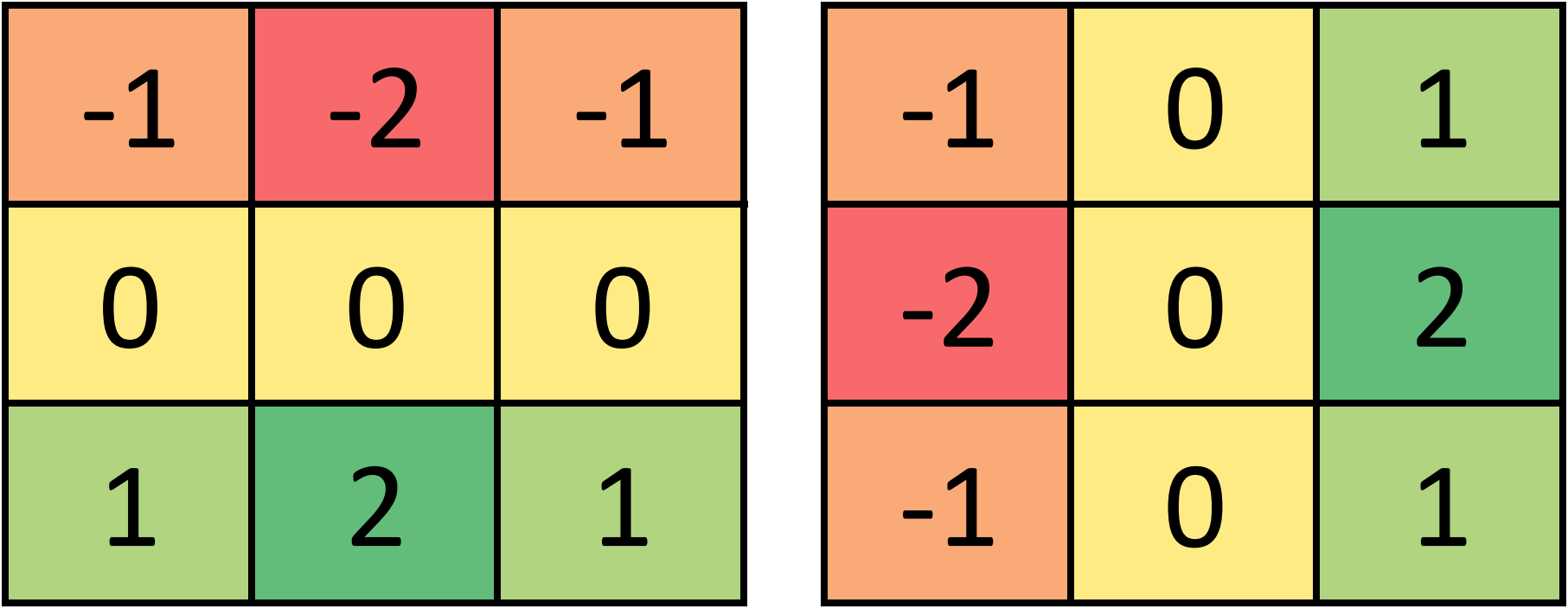} 
\caption{Sobel 3$\times$3 mask}
\label{fig:sobel}
\end{center}
\end{figure}

The earliest operator that was used in the field of edge detection was Robert's cross-gradient operator \cite{gonzalez2009digital}. It computes the gradient of an image by using 2-D masks and gives preference to diagonal edges. In order to be symmetric about the center point, the smallest metric should be of  3$\times$3 dimensions. The extension of 2D masks to 3D masks gives a new operator known as the Prewitt’s Operator\cite{gonzalez2009digital}. Sobel Operator, shown in Fig.~\ref{fig:sobel} is a modified form of Prewitt's operator in which the central column/row is multiplied by a factor of 2. This is equivalent to combining the  smoothing operation with Prewitt method. Segmentation of images through edge-based approaches have certain disadvantages such as contour filling, need of optimal value of threshold for true edges, and the requirement of post-processing.


\subsection{Watershed Segmentation}

One of the popular approaches in this category is based on the concept of morphological watersheds. In watershed segmentation \cite{beucher1992morphological,belaid2011image}, an image is visualized in three dimensions with (x,y) coordinates on $x$ and $y$ axes and intensity on the $z$ axis.
This approach  models an image as a topographical plane in which pixel intensities are equivalent to the height of mountains. Each intensity level is represented by a plane. The topographic interpretation divides all points of image into three categories: regional minimum, catchment basin and watershed lines. The \textit{catchment basin} includes those points at which if a drop of water is placed, the drop would always fall  to a single regional minimum. The \textit{watershed lines} contain those points at which water droplet can fall to more than one regional minimum. 

\begin{figure*}[t]
	
	\begin{subfigure}{.488\textwidth}
		\centering
		\includegraphics[width=0.8\linewidth]{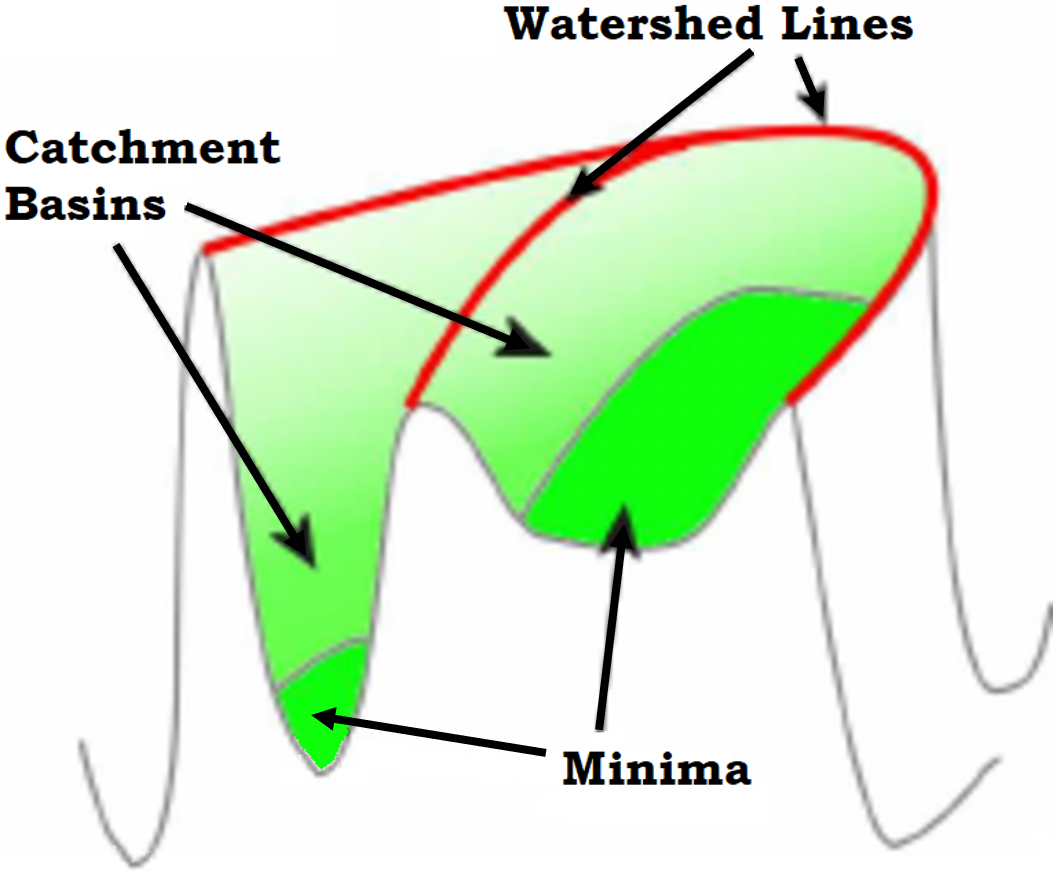}
		\caption{}
		\label{fig:catchment}
	\end{subfigure}%
	\hfill
	\begin{subfigure}{.488\textwidth}
		\centering
		\includegraphics[width=0.8\linewidth]{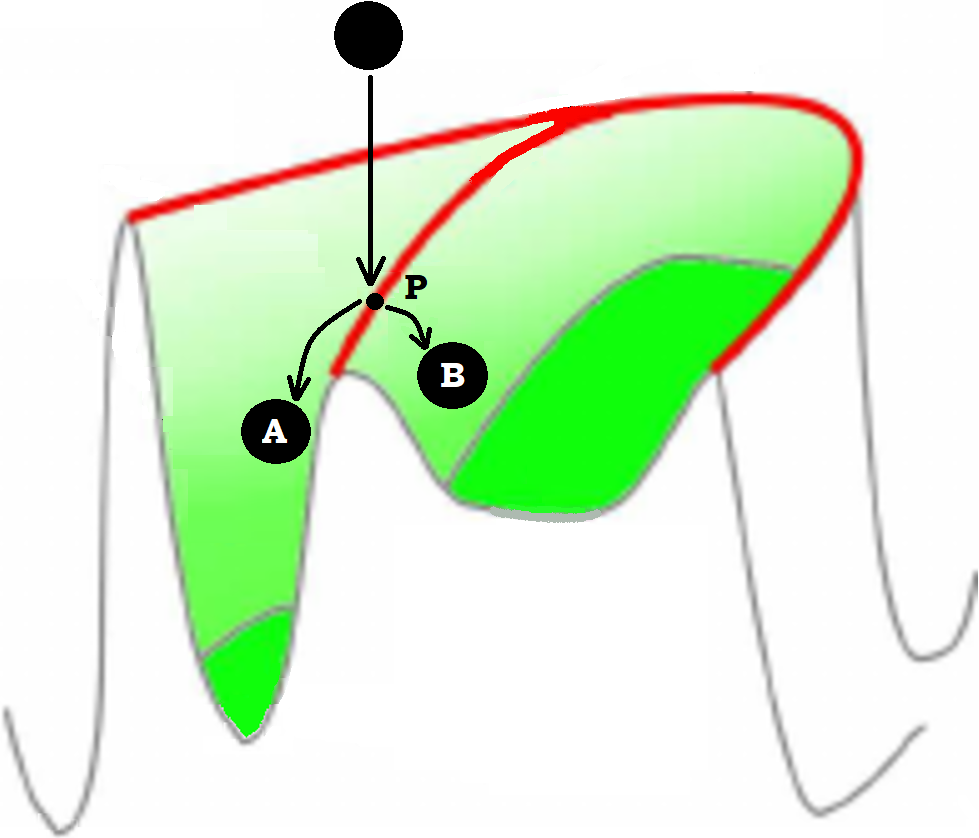}
		\caption{}
		\label{fig:falling_ball}
	\end{subfigure}
	
	\caption{(a) Catchment basins \& watershed lines and (b) a ball falling from hill, and striking the watershed line at point `P'  will go to either `A' \textbf{or} `B'}
	\label{fig:ball_catchment}
\end{figure*}

The basic idea of watershed algorithm is that holes are punched in every regional minimum, and water is allowed to rise through holes at a uniform rate. Here, water is flooded from minimum intensity to maximum intensity of image. When the rising water level  in one catchment basin is about to merge with water from another catchment basin, a dam is built between them to prevent merging. These dams are those points which are the results of watershed segmentation and they depict the separation lines of object from the background image. An illustration of the catchment basins and the watershed lines is provided in Fig.~\ref{fig:catchment}.

Let the $M_{1}, M_{2}, ..., M_{R}$ be regional minimum of an image $ g(x,y) $. Let T[n] denote the set of all those points which lie below the plane $n$, where value of $n$ varies between minimum to maximum intensity in the image. That is:
\begin{equation}
T[n]=\{(s,t)|g(s,t)<n\}
\end{equation}
Let $C_{n}(M_{i})$ denote the set of points in the catchment basin associated with regional minimum $M_{i}$ that are flooded at plane $n$. It can be computed by
\begin{equation}
C_{n}(M_{i})=C(M_{i})\cap T[n]
\end{equation}
where, $C(M_{i})$ is the set of points in the catchment basin associated with regional minimum $ M_{i} $.

Finally $C[n]$ denotes the union of all flooded catchment basins at particular stage $n$:
\begin{equation}
C[n]=\cup C_{n}(M_{i})
\end{equation}
The algorithm computes $C[n]$ from $C[n-1]$ recursively and it initializes with $C[min+1]=T[min+1]$. It first computes the set $Q$ for all connected components of $T[n]$. Thereafter, it takes each component $q~{\in}~Q[n]$ one by one, and checks amongst the following three possibilities:
\begin{itemize}
	\item $ q\cup C[n-1] $ is empty. This condition occurs when a new minimum is encountered.
	\item $ q\cup C[n-1] $ contains one connected component of $ C[n-1] $. It occurs when $ q $ lies within the catchment basin of some regional minimum.
	\item $ q\cup C[n-1] $ contains more than one connected component of $ C[n-1] $. It occurs when all, or a part of one or more ridges separating two or more catchment basins are encountered. In this case, there are chances of overflows. It is in this situation that a dam must be constructed between the catchment basins.
\end{itemize}

Suppose that at stage $ n-1 $, we have two separate catchment basins which can be merged at stage $ n $ through flooding. In order to prevent overflow, we construct a dam between these catchment basins. Let the connected component be $q$ and two separated components from step $n-1$ are $ q\cap C[n-1] $. For the construction of dam, the separated components are dilated by the structuring element. The dilation is subjected to two conditions: (1) The center of the structuring element can be located only at points in $q$. (2) The dilation cannot be performed on points that would cause the sets being dilated to merge.

There are some problems in watershed segmentation, such as the dependence  on initial consideration of catchment basins, high computational complexity, and over-sensitivity for complex images. Therefore, as shall be shown later in this paper, watershed segmentation is not satisfactory for complex biomedical images.

From the above discussion, it may be noted that although edge detection algorithms are simple, but when used for segmentation purposes, they do not guarantee closed contours and are sensitive to threshold selection. On the other hand, watershed transform gives closed contours, but is over sensitive for complex images. Therefore a new algorithm is proposed by utilizing the concepts of watershed and edge detection methods.

\subsection{Fuzzy Multi-Criteria Decision Making}

Fuzzy theory is a powerful tool to represent some form of uncertainty or vagueness \cite{kaushik2011artificial}, where the uncertainty may be attributed to partial or unreliable or conflicting information. Unlike classical theory, fuzzy theory allow us to represent information up to a certain degree, i.e. not only in terms of true(1) or false(0) but also in terms of partially true, partially false etc. This implies that values in  fuzzy logic can be any number between 0 and 1 (both inclusive).

A membership function can be defined as a curve that indicates how each point in an input space is mapped to a membership value between 0 and 1. The membership function can be represented by a curve whose shape defines a function that suits us from the point of view of simplicity, convenience, speed, \& efficiency, and must satisfy the condition that it varies between 0 and 1. The membership function is generally represented by $ \mu $ that maps each element of $ \cup $ (Universe of discourse). Some  examples of membership functions are: Triangular Membership Function, Trapezoidal Membership Function, Gaussian Distribution Function, Generalized Bell Membership Function, Sigmoidal Membership Function, Polynomial-based Curves, and Vicinity Function.

Multiple Criteria Decision Making \cite{kahraman2008fuzzy} was introduced in the early 1970’s. Multi-Criteria Decision Making (MCDM) is a tool that we can apply in taking complex decisions, i.e. in those situations in which decision depends on more than one criterion. For example in everyday life, a typical example of MCDM is buying a laptop, wherein the different criteria would be price, brand, processor, memory, size, color, weight, etc. Therefore, by considering all these criteria with different weightages, a decision will be taken.

Bellman and Zadeh in 1970 \cite{bellman1970decision} and Zimmermann in 1978 \cite{zimmermann1978fuzzy}, were amongst the first to introduce fuzzy sets into MCDM. Both these works consider the space in which the different constraints and goals are defined, and redefine (in the same space) fuzzy goals and fuzzy constraints symmetrically as fuzzy sets. Then the decision of maximization was taken at the point in the space at which membership function attained its maximum value. Fuzzy Multi-criteria Decision Making is becoming a powerful tool for many real-life applications. A brief overview of categories of MCDM and its applications is contained in \cite{abdullah2013fuzzy}. They find many of its applications in consumer purchasing, biotechnology, road accident problems, sports etc.

Fuzzy image processing \cite{kerre2000fuzzy,chi1996fuzzy} has also become an emerging branch of research in the present time, because of its many benefits in solving image problems. It is known for its capability of managing vagueness and ambiguity efficiently. On the other hand, data of image is also random and many difficulties arise in image processing because of such uncertainty. So fuzzy theory can prove to be a good technique for solving image processing problems like image segmentation.

\section{The Proposed Falling-Ball Algorithm}
\label{sec:proposed}

A novel algorithm for image segmentation, based on Fuzzy MCDM, is proposed. It is named as the \textit{Falling-Ball Algorithm}. The output of this algorithm is the segmented image.  The Falling-Ball algorithm uses some concepts of watershed image segmentation algorithm. For instance, it works by using the topographic view of image, in which height at a particular point is proportional to intensity of the pixel at the corresponding point in the image. It finds edges using catchment basins, but unlike the watershed algorithm, in the Falling-Ball algorithm neither dilation nor dam construction takes place. Also the procedure of finding catchment basins is different from watershed segmentation. The catchment basins includes those areas, which are enclosed from surroundings by high-rise areas e.g. hills. An illustration of the catchment basins is shown in Fig.~\ref{fig:catchment}.

The Falling-Ball Algorithm uses the concept of an imaginary ball which is allowed to fall down from every pixel of an image. This is different from the watershed segmentation technique where the topography is flooded with water from bottom. So, if a water droplet is placed on a point of hill where it has more than one option (direction) to fall, then it may divide itself into two parts to fall. However, a ball is different from a water droplet in the sense that it cannot break, and therefore can fall only in one direction. This leads to different membership functions for different options. It can be seen in Fig.~\ref{fig:falling_ball} that a ball dropped towards point `P' can either go to `A' \textit{or} towards `B', and can only fall in any one of the reachable catchment basin. From this step, the algorithm finds the center of different catchment basins. Then it expands those catchment basins by retracing the path of ball. In the process of retracing, it also finds the membership value of each pixel corresponding to each catchment basin.

The membership function used here depends on \textit{Multi-Criteria Decision Making}, details of which are provided along with the explanation of the algorithm in Section~\ref{sec:implementation}. As discussed earlier, after the execution of the Falling-Ball algorithm, all pixels get their membership values corresponding to each catchment basin. Thereafter, it is required to design a \textit{classifier} to use these membership values and assign different pixels to their correct catchment basins. This is done in order to ensure that those pixels, for which decision of assignment to one catchment basin is not defined, can be identified as edge pixels. If all membership values are crisp instead of fuzzy, then the solution will be trivial, each pixel can be assigned to that catchment basin whose corresponding membership value with that pixel is 1. But in such a case, even though both the algorithm and the classifier are simpler,  it will not give any edge pixel. This is because all pixel membership vectors contain only values 0 and 1, \textit{i.e.} they either belong or do not belong to a catchment basin. This is the primary reason of the use of fuzzy theory with multi-criteria decision making, so that it can take all criteria and give a meaningful decision by considering all factors.

In the construction of the fuzzy classifier, we first find the threshold of each catchment basin. Thereafter, we consider the membership value vector corresponding to each pixel in the image and find the two largest values in that vector. Catchment basins, and their thresholds, are then identified for these two largest values from the membership value vector.
If any one value (out of  two largest values found) has a value greater than the thresholds of  catchment basins (identified in the previous step), then the pixel under consideration is identified to be of that particular catchment basin. Details of the classifier are included with the discussion of the algorithm in the next section. After having identified the edges by the above mentioned process, a closed contour algorithm is  applied on the edges to form closed regions. Lastly, an object separation procedure is applied on the resultant image to get the finally segmented object(s).

\section{Implementation Algorithm}
\label{sec:implementation}

\begin{figure}[!ht]
\begin{center}
\includegraphics[width=0.49\textwidth]{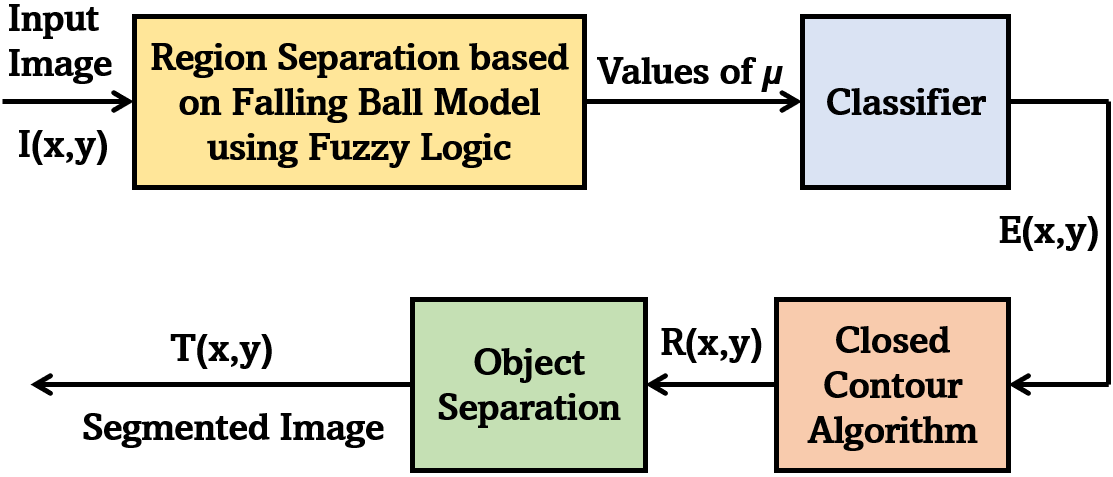} 
\caption{Block diagram of the proposed Falling-Ball algorithm}
\label{fig:flow}
\end{center}
\end{figure}

The Falling-Ball algorithm is based on automatic threshold calculation and uses the following four steps: \begin{enumerate}
	\item Construct catchment basins and find values of $\mu$
	\item Classify edge pixels using a classifier
	\item Apply Closed-Contour algorithm
	\item Object segmentation based on pixel intensity within the closed contour.
\end{enumerate}
The block diagram of the above approach is shown in Fig.~\ref{fig:flow}.
\subsection{Construction of catchment basins and values of $\mu$}

Consider the original image as a  two-dimensional matrix $I(i,j)$ with $h$ rows and $w$ columns, where $h$ and $w$ are height and width of image respectively. The catchment basins are denoted by $ C_{1}, C_{2}, ..., C_{k} $, where $k$ is the number of catchment basins, which is initialized to 0. The following algorithm finds the center $ (i_{c},j_{c}) $ of each catchment basin by calling \textit{Ball-Fall} procedure and explores this catchment basin by finding more member pixels by calling the \textit{Expand} procedure. Similarly, it goes on to identify all catchment basins.
\begin{enumerate}
	\item for $i =$ 1 to $h$
	\item ~~~~for $j =$ 1 to $w$
	\item ~~~~~~~~\{ $ (i_{c},j_{c}) \leftarrow$ Ball-Fall$(i,j)$
	\item ~~~~~~~~~~~$k=k+1$ //increment counter of catchment basin
	\item ~~~~~~~~~~~$\mu[i_{c}][j_{c}][k] \leftarrow 1$
	\item ~~~~~~~~~~~call Expand($i_{c},j_{c}$)
	\item ~~~~~~~~\}\\
\end{enumerate}

The Ball-Fall procedure allows a hypothetical ball to fall from pixel $ i_{0},j_{0} $ until the energy of ball becomes zero. In other words, the ball is falling from top of hill to the plane and then stopping when the constant plane ends, and another upward slope starts. It returns this point as center of catchment basin.\\

\textbf{Ball-Fall($ i_{0},j_{0} $)}
\begin{enumerate}
	\item for $n = $ 1 to 8 neighbors of pixel($ i_{0},j_{0} $)
	\item ~~~~\{~~~~if($ I(i_{0},j_{0})\leq I(i_{n},j_{n}) $)
	\item ~~~~~~~~~~\{~~~~Ball-Fall($ i_{n},j_{n} $)
	\item ~~~~~~~~~~\}
	\item ~~~~~~~else\{~~~~$ i_{c} \leftarrow i_{n} $ and $ j_{c}\leftarrow j_{n} $
	\item ~~~~~~~~~~~~~~~~~return
	\item ~~~~~~~~~~~\}
	\item ~~~~\}\\
\end{enumerate}

The Expand procedure scans all descendants, and if their intensity is higher than the base pixel intensity, then the membership values for the neighbours are found. The procedure is recursively repeated for all pixels of each catchment basin. Consider \textit{diff}, \textit{x}, \textit{t} and \textit{n} as different variables and \textit{temp} as variable vector. \\

\textbf{Expand($ i_{0},j_{0} $)}
\begin{enumerate}
	\item for $n =$ 1 to 8 neighbors of pixel($ i_{0},j_{0} $)
	\item ~~~~\{~~~~if($ I(i_{n},j_{n})\geq I(i_{0},j_{0}) $)
	\item ~~~~~~~~~~\{~~~~$ diff \leftarrow I(i_{n},j_{n}) - I(i_{0},j_{0}) $
	\item ~~~~~~~~~~~~~~~~$ x \leftarrow (255-diff)/255 $
	\item ~~~~~~~~~~~~~~~~for $t =$ 1 to k
	\item ~~~~~~~~~~~~~~~~\{~~~$temp[t] \leftarrow \mu[i_{0}][j_{0}][t]*x$
	\item ~~~~~~~~~~~~~~~~\}
	\item ~~~~~~~~~~~~~~~~for $t =$ 1 to k
	\item ~~~~~~~~~~~~~~~~\{~~~if($ \mu[i_{n}][j_{n}] < temp[t] $)
	\item ~~~~~~~~~~~~~~~~~~~~~~~~~~~$ \mu[i_{n}][j_{n}]=temp[t] $
	\item ~~~~~~~~~~~~~~~~\}
	\item ~~~~~~~~~~~~~~~~call expand($ i_{n},j_{n} $)
	\item ~~~~~~~~~~\}
	\item ~~~~\}
\end{enumerate}

\subsection{Classify edge pixels using a classifier}

The membership function ($ \mu $) values for all pixels are already computed for all catchment basins ($ C_{1},C_{2},...C_{k} $) using the above procedures, i.e.

\begin{center}
$ \mu[0][0] = [\mu_{1_{0,0}}, \mu_{2_{0,0}}, ..., \mu_{k_{0,0}}] $\\
$ \mu[0][1] = [\mu_{1_{0,1}}, \mu_{2_{0,1}}, ..., \mu_{k_{0,1}}] $\\
.\\.\\.\\
$ \mu[i][j] = [\mu_{1_{i,j}}, \mu_{2_{i,j}}, ..., \mu_{k_{i,j}}] $\\
.\\.\\.\\
$ \mu[h][w] = [\mu_{1_{h,w}}, \mu_{2_{h,w}}, ..., \mu_{k_{h,w}}] $\\  
\end{center}

In the classifier, first a new image is created with all pixels being white (initialized to RGB = `FFFFFF'). The threshold values ($ Th_{1}, Th_{2}, ..., Th_{k} $) are then calculated for each catchment basin, where $k$ is the number of catchment basins. For each membership function vector ($ \mu[0][0], \mu[0][1]...\mu[i][j], ..., \mu[h][w] $), we find the two largest values ($L_{1} \&  L_{2}$), and then determine the catchment basin  $ k_{1} $ corresponding to largest value of $ \mu[i][j] $ and $ k_{2} $ for second largest value of $ \mu[i][j] $. If the difference between the $Th_{k_{1}}$ and $ L_{1} $ is less than 0.11 (i.e. 11\%), then we make that pixel black in the initialized white image, otherwise check for $ L_{2} $. If the difference between the $Th_{k_{2}}$ and $ L_{2} $ is less than 0.11, then we  make the pixel black. At the end, all pixels which remain white after this process is completed denote the  edges. The algorithm is described as follows:\\
\begin{enumerate}
	\item for i=0 to h
	\item ~~~~for j=0 to w
	\item ~~~~~~~~$E[i][j] \leftarrow 255$
	\item for t=1 to k
	\item ~~~~$ Th_{t} \leftarrow max \left\lbrace \mu[0][0][t],\mu[0][1][t],....\mu[h][w][t] \right\rbrace $
	\item for i=0 to h
	\item ~~~for j=0 to w
	\item ~~~\{
	\item ~~~~~~$L_{1} \leftarrow first\_max \left\lbrace \mu[i][j][0],\mu[i][j][1],....\mu[i][j][k] \right\rbrace $
	\item ~~~~~~~~~~for t=0 to k
	\item ~~~~~~~~~~\{~~~if($L_{1}=\mu[i][j][t]$)
	\item ~~~~~~~~~~~~~~~\{~~~$ k_{1}=t $
	\item ~~~~~~~~~~~~~~~~~~~~break
	\item ~~~~~~~~~~~~~~~\}
	\item ~~~~~~~~~~\}
	\item ~~~~~~$ L_{2} \leftarrow second\_max \left\lbrace \mu[i][j][0],\mu[i][j][1],....\mu[i][j][k] \right\rbrace $
	\item ~~~~~~~~~~for t=0 to k
	\item ~~~~~~~~~~\{~~~if($L_{2}=\mu[i][j][t]$)
	\item ~~~~~~~~~~~~~~\{~~~$ k_{2}=t $
	\item ~~~~~~~~~~~~~~~~~~break
	\item ~~~~~~~~~~~~~~~\}
	\item ~~~~~~~~~~\}
	\item ~~~~~~~~~~if ($ |Th_{k_{1}}-L_{1}|<0.11 $)
	\item ~~~~~~~~~~~~~~$ E[i][j]=0 $
	\item ~~~~~~~~~~if ($ |Th_{k_{2}}-L_{2}|<0.11 $)
	\item ~~~~~~~~~~~~~~$ E[i][j]=0 $
	\item ~~~~\}\\
\end{enumerate}
Here, $first\_max$ and $second\_max$ are functions in which $first\_max$ returns the first largest value in the vector, while $second\_max$ returns the second largest value in the vector.

\begin{figure}[!h]
	\centering
	\begin{subfigure}{.125\textwidth}
		\centering
		\begin{center}
			\includegraphics[width=.8\linewidth]{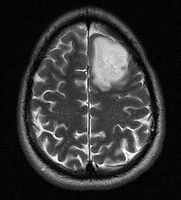}
			\caption{Brain-Tumor}
		\end{center}
		\label{fig:sfig1}
	\end{subfigure}%
	
	\begin{subfigure}{.122\textwidth}
		\centering
		\includegraphics[width=.8\linewidth]{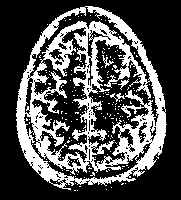}
		\caption{constant 0.05}
		\label{fig:sfig2}
	\end{subfigure}%
	\hfill
	\begin{subfigure}{.122\textwidth}
		\centering
		\includegraphics[width=.8\linewidth]{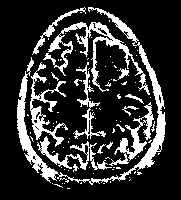}
		\caption{constant 0.1}
		\label{fig:sfig2}
	\end{subfigure}%
	\hfill
	\begin{subfigure}{.122\textwidth}
		\centering
		\includegraphics[width=.8\linewidth]{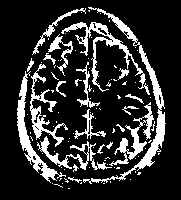}
		\caption{constant 0.11}
		\label{fig:sfig2}
	\end{subfigure}%
	\hfill
	\begin{subfigure}{.122\textwidth}
		\centering
		\includegraphics[width=.8\linewidth]{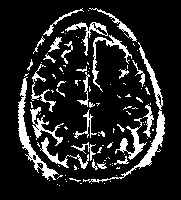}
		\caption{constant 0.5}
		\label{fig:sfig2}
	\end{subfigure}
	
	\caption{Reason for Threshold Constant 0.1}
	\label{fig:const_fig}
\end{figure}

Prior to a discussion of the closed contour algorithm, it is necessary to give the reason for choosing the threshold. It is mentioned in the algorithm that we are giving a relaxation of 11\% in threshold. For finding this percentage, we performed experiments on a set of images and found that value as the most suitable constant for getting superior results. The test results for the experiments to find the threshold, for one sample image,  are shown in Fig.~\ref{fig:const_fig}, for different values of thresholds. It can be seen that smaller values lead to the presence of false edges, while larger thresholds result in the elimination of true edges. Therefore, the value of 11\% is taken as an optimal value of threshold.

The closed contour algorithm is then applied to the output of classifier. The regions are formed from the edges detected in the image which is the output of the classifier. The region of interest is separated here, after the detection of all closed contours. The brain tumor is extracted from the `regions' image by using the following property of brain tumor images:  it is known that brain tumor is an active region of the MRI scan of brain and the active region can be recognized by brighter color areas.

\subsection{Closed Contour Algorithm}

\begin{figure}[hb]
\begin{center}
\includegraphics[width=0.33\textwidth]{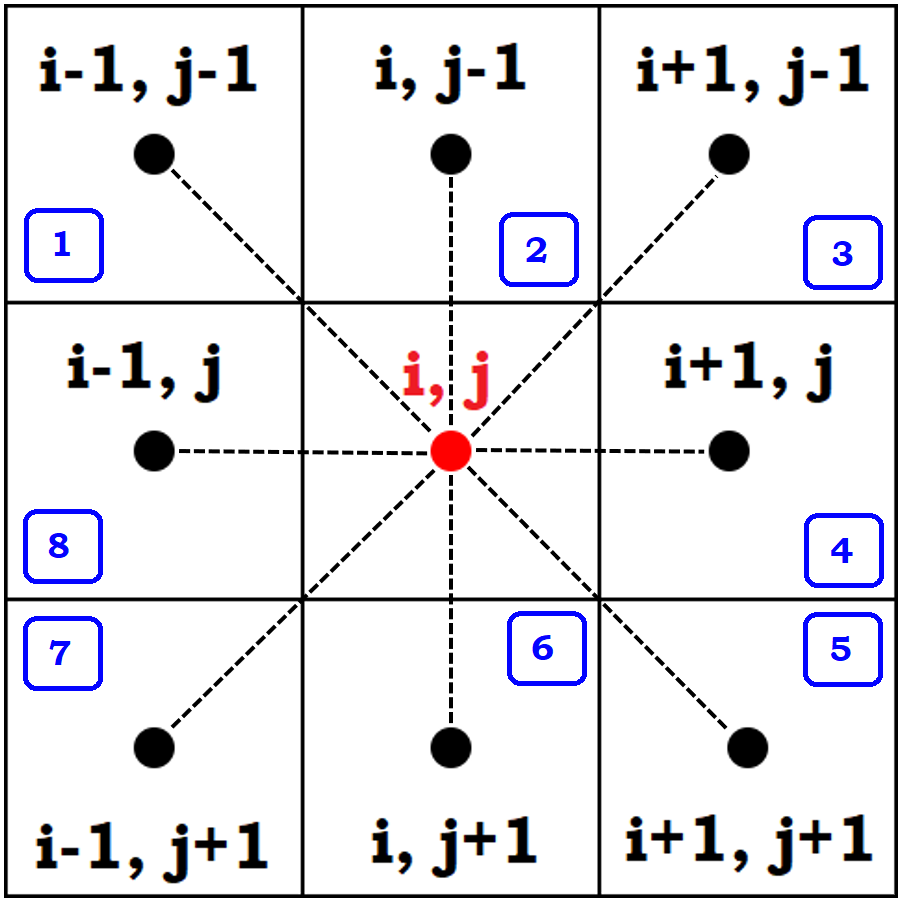} 
\caption{8-connected neighbour}
\label{fig:8}
\end{center}
\end{figure}

The closed contour algorithm for the Falling-Ball method first finds a seed pixel of each region, then it expands that region with the help of its 8-neighbours as shown in Fig~\ref{fig:8}. It scans the image pixel by pixel, to first find the pixel that can be inside a region (if it does not lie on any edge). Then the algorithm checks for 8-connected neighbours; i.e. if they belong to this region or not (that is, if they are not a member of other regions and are not the edge pixels), then they can be the part of this region. The same process is repeated iteratively for neighbours of neighbours and so on. Therefore, it is a recursive procedure. Meanwhile it also checks for the boundary of other regions by considering a 5$\times$5 window around every pixel, and it stops if any pixel in the window is boundary pixel. Although it gives a slightly shrunk region because it finds closed contours 2 pixels away from the boundary of regions, it has the advantage of finding closed contours in all cases. The proposed algorithm is explained next. 

Consider the image, obtained after thresholding, as a two-dimensional matrix $ E(i,j) $ with `\textit{h}' rows and `\textit{w}' columns. The algorithm scans $ E(i,j) $ and segments it into $ r $ regions. Thereafter, it associates each pixel with one of the $r$ regions. Let $ R_{k} $ denote the $ k^{th} $ region and initialize $ r $ with 0. 

\begin{enumerate}
\item for i=1 to h
\item ~~~~for j=1 to w
\item ~~~~~~~~if E(i,j)=0
\item ~~~~~~~~\{~~~~for k=1 to r
\item ~~~~~~~~~~~~~~~~if( E(i,j) $ \epsilon $ region $ R_{k} $)
\item ~~~~~~~~~~~~~~~~~~~~ \{ goto step 1 to scan next pixel
\item ~~~~~~~~~~~~~~~~~~~~ \}
\item ~~~~~~~~~~~~r=r+1~~~~//increment the region index by 1
\item ~~~~~~~~~~~~E($ i,j $) $\rightarrow$ $ R_{r} $~//insert element in region $ R_{r+1} $
\item ~~~~~~~~~~~~call search(i,j)
\item ~~~~~~~~\}\\
\end{enumerate}
Now after the first pixel of a particular region $R_{r}$ is known,   the following search procedure is called to find more pixels of this region: \\

\textbf{Search($ i_{0},j_{0} $)}
\begin{enumerate}
\item for n=1 to 8 neighbours of pixel($ E(i_{0},j_{0}) $)
\item ~~~~\{ for $k=1$ to r
\item ~~~~~~~~\{ if($ E(i_{n},j_{n}) $ $ \epsilon $ region $ R_{k} $)
\item ~~~~~~~~~~~~\{ goto step 1 to scan next neighbour
\item ~~~~~~~~~~~~\}
\item ~~~~~~~~\}
\item ~~~~for $x=-2$ to $2$
\item ~~~~~~~~for $y=-2$ to $2$
\item ~~~~~~~~~~~~if($E(i_{n}+x,j_{n}+y)=255$)~~~goto step 1
\item ~~~~E($i_{n},j_{n}$) $\rightarrow $ $ R_{r} $ ~~~~//insert element in region $ R_{r} $
\item ~~~~call search($i_{n},j_{n}$)
\item ~~~~\}
\end{enumerate}

\subsection{Object Separation}

It is expected that after the previous three steps, all closed contours are detected. The next step is the extraction of tumor from the input image. The brain tumor is extracted from the `regions' image $ R(x,y) $ by detecting the most active region of the brain MRI. Active region is recognized by brighter color area. The following equation is used for locating the tumor.
\begin{equation}
Tumor=R_{k}|\eta_{k}=max \left\lbrace \eta_{1}, \eta_{2}, ..., \eta_{r} \right\rbrace
\end{equation}
where, 
$ R_{i} $ denotes $ i_{th} $ region, and 
$ \eta_{i} $ denotes average intensity of pixels of MRI that overlap with region $ R_{i} $.

Consider the image generated after separating the regions be $ R(x,y) $ and original image is $ f(x,y) $. From these, we create a new image $ T(x,y) $ of the tumor. Let $ R_{k} $ denote the $ k^{th} $ region and $ r $ be the total number of regions in $ R(x,y) $,  then the algorithm can be explained as follows:

\begin{enumerate}
\item for t = 1 to r
\item ~~~~\{
\item ~~~~~~~$ S_{t} \leftarrow 0 $ ~~~~//set sum $ S_{t}=0$	
\item ~~~~~~~$ n_{t} \leftarrow 0 $ ~~~~//set number of pixels $ n_{t}=0$	
\item ~~~~~~~for i=1 to h
\item ~~~~~~~~~~~for j=1 to w
\item ~~~~~~~~~~~~~~~if ( pixels $(i,j) \epsilon  R_{t}$)
\item ~~~~~~~~~~~~~~~~~~~$ S_{t}=S_{t}+f(x,y) $
\item ~~~~~~~~~~~~~~~~~~~$ n_{t}=n_{t}+1 $
\item ~~~~~~~$ \eta_{t}=S_{t}/n_{t} $
\item ~~~~ \}
\item $ \eta_{k}max \left\lbrace \eta_{1},\eta_{2},...\eta_{r} \right\rbrace $
\item for i=1 to h
\item ~~~~for j=1 to w
\item ~~~~~~~~T(i,j) $\leftarrow$ 0
\item ~~~~~~~~if ( pixels $(i,j) \epsilon  R_{k}$)
\item ~~~~~~~~~~~~$ T(i,j)=f(i,j) $
\end{enumerate}

\section{Simulation Results}
\label{sec:comparison}

\begin{figure}[tb]
	\begin{center}
		\includegraphics[width=0.499\textwidth]{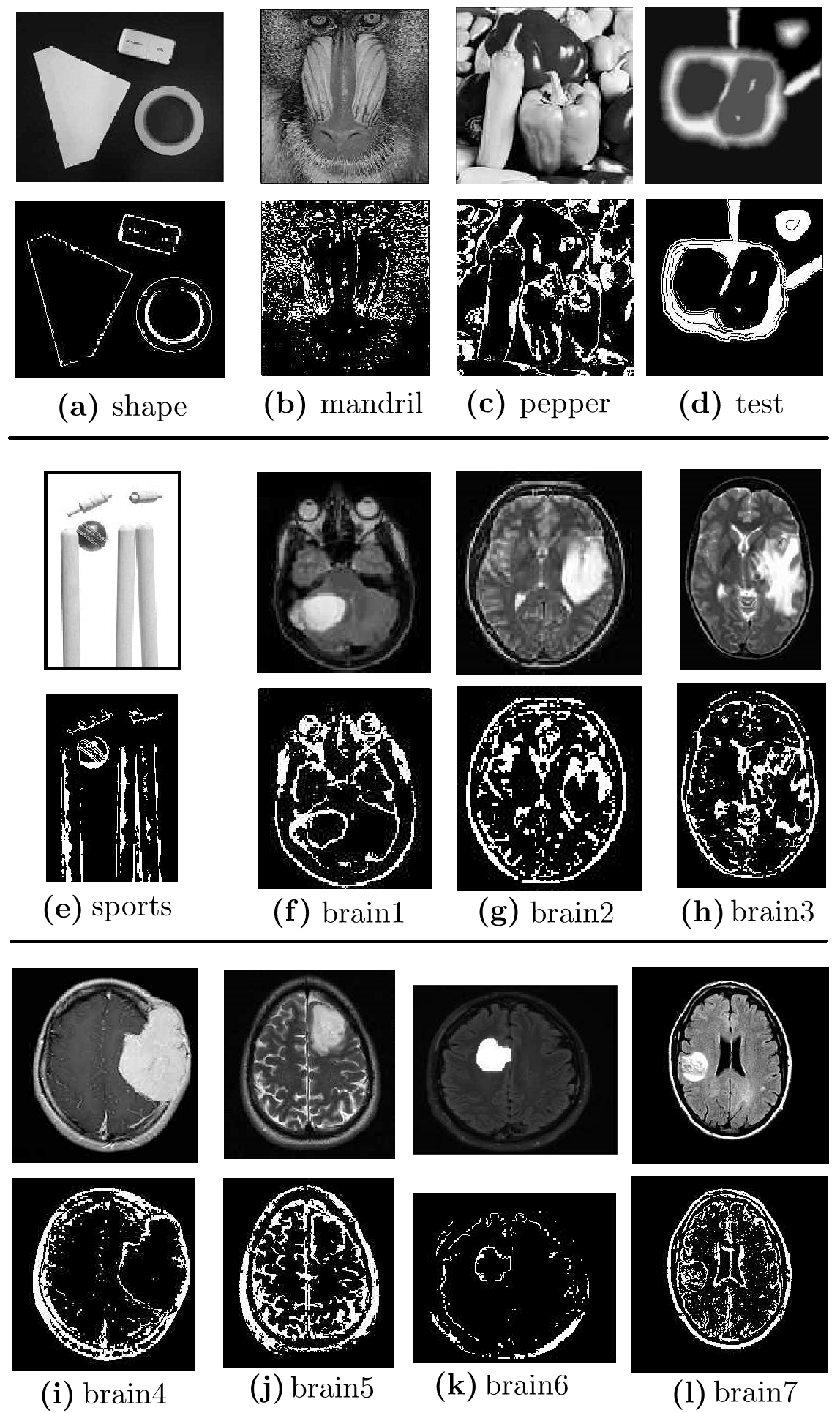}
		\caption{Edge detection results using the proposed Falling-Ball Algorithm} \label{fig:sample_results}
	\end{center}
\end{figure}

Results of the application of the proposed algorithm to different images are presented in Fig.~\ref{fig:sample_results}. The edges detected by the Falling-Ball algorithm after the steps of the application of the falling ball and the classifier are shown. The edges in \textit{(a) shape} image are at the boundary of objects with some absence of true edges. In the \textit{(b) mandril} image, edges are very clear and many false edges are absent. Edges in the \textit{(c) pepper} image are also visible clearly with fine lines for all objects. Same is the case with another shown test image i.e. \textit{(d) test}. The edges of the stumps detected in \textit{(e) sports}  
by Falling-Ball algorithm are shown. Here, both edges of each stump have been correctly detected. Thereafter, samples of the algorithm test results over several brain tumor images are presented in Fig.~\ref{fig:sample_results}(f) through Fig.~\ref{fig:sample_results}(l).

Subjective as well as numerical comparisons are next performed to demonstrate the superiority of proposed method over the watershed edge detection method, in identifying the brain tumors. For numerical comparison, three parameters as defined below, are considered.
\begin{enumerate}
\item
Gray level Uniformity measure (GU)\cite{borsotti1998quantitative}:\\
The gray level uniformity measure is based on inter-region uniformity. The uniformity of a feature over a region can be computed on the basis of the variance evaluated at every pixel belonging to that region. For an image $f(x,y)$, the gray level uniformity measure (GU) can be computed as:
\begin{equation}
GU=\sum_i \sum_{(x,y)\epsilon R_{i}} \left[f(x,y)-\frac{1}{A_{i}}\sum_{(x,y)\epsilon R_{i}} f(x,y)\right]^{2}
\label{method1}
\end{equation}
Lower value of GU is desirable.\\
\item
Q-parameter\cite{zhang1996survey}:\\
The Q-parameter is based on three criteria: \textit{(i)} region must be uniform; \textit{(ii)} region's interior should not have too many holes; \textit{(iii)} adjacent regions must have non-uniform characteristics. The function $Q(I)$ for performing this task is:
\begin{small}
\begin{equation}
Q(I)=\frac{1}{1000(N*M)}\sqrt{R}*\sum_{i=1}^R \left[\frac{e_{i}^{2}}{1+logA_{i}}+\left(\frac{R(A_{i})}{A_{i}}\right)^{2}\right]
\label{method2}
\end{equation}
\end{small}
Lower values of Q(I) mean better performance.\\ 
\item
Relative Ultimate Measurement Accuracy (RUMA)\cite{borsotti1998quantitative}:\\
The Relative Ultimate Measurement Accuracy (RUMA) is defined as:
\begin{equation}
RUMA=\frac{|R_{f}-S_{f}|}{R_{f}}*100
\label{method3}
\end{equation}
where $R_{f}$ denotes the feature value obtained from the reference image and $S_{f}$ denotes the feature value measured from the segmented image. Lower the value of RUMA, better is the performance of segmentation algorithm.\\
\end{enumerate}

\begin{figure}[tb]
	\begin{center}
		\includegraphics[width=0.47\textwidth]{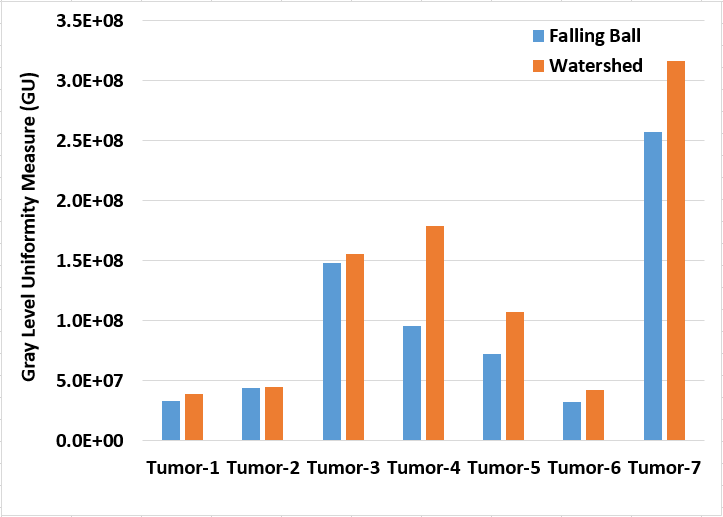}
		\caption{Performance comparison in terms of Gray level Uniformity measure(GU)} \label{fig:bar-chart-method-1}
	\end{center}
\end{figure}

\begin{figure}[tb]
	\begin{center}
		\includegraphics[width=0.47\textwidth]{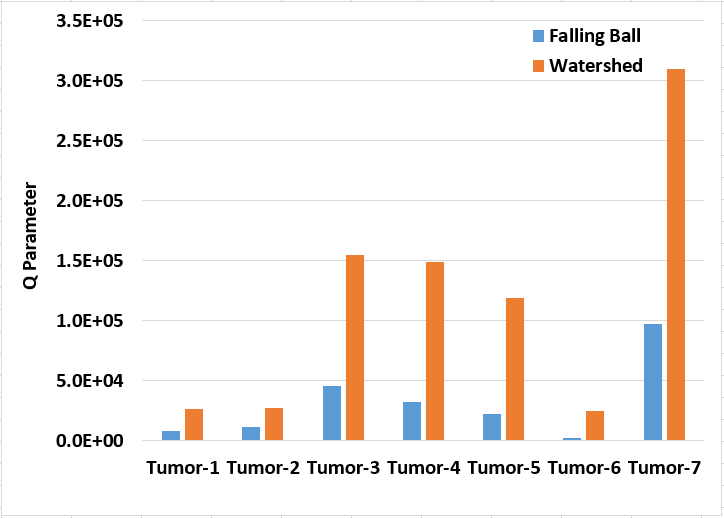}
		\caption{Performance comparison in terms of Q-parameter} \label{fig:bar-chart-method-2}
	\end{center}
\end{figure}

\begin{figure}[tb]
	\begin{center}
		\includegraphics[width=0.47\textwidth]{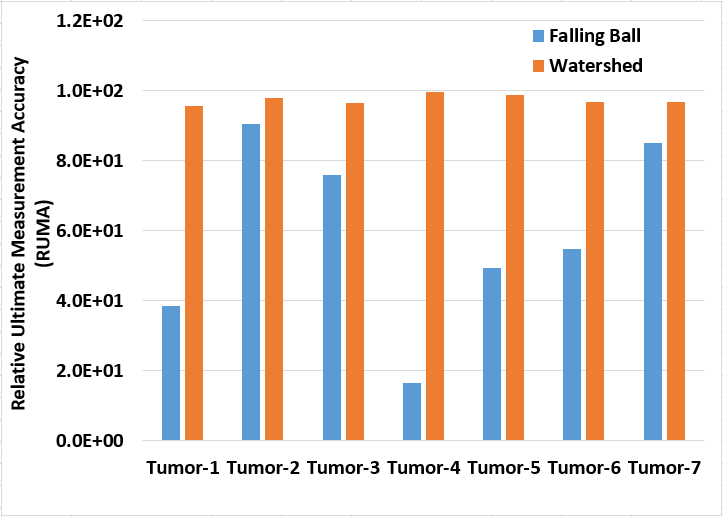}
		\caption{Performance comparison in terms of Relative Ultimate Measurement Accuracy(RUMA)} \label{fig:bar-chart-method-3}
	\end{center}
\end{figure}

The performance of the proposed algorithm is evaluated for seven MRI images (Tumor 1 to 7) in terms of GU, Q and RUMA, and is shown in Fig. \ref{fig:bar-chart-method-1}, \ref{fig:bar-chart-method-2} and \ref{fig:bar-chart-method-3} respectively. For comparison, the values of these parameters are also evaluated for watershed segmentation. It can be observed from Fig.~\ref{fig:bar-chart-method-1} that for almost all tumor images, the values of GU using the proposed algorithm are lower than the GU values obtained with watershed algorithm. 
%
%
Similarly from Fig.~\ref{fig:bar-chart-method-2}, it can be seen that the Q-parameter values for the proposed method are much lower than the Q-values obtained with watershed method,  for all the seven tumor images.
Lastly, from Fig.~\ref{fig:bar-chart-method-3}, it can be observed that RUMA values obtained from proposed algorithm are much lower than corresponding values of watershed edge detection for all images.
\begin{figure*}[tb]
	\begin{center}
		\includegraphics[width=0.7\textwidth]{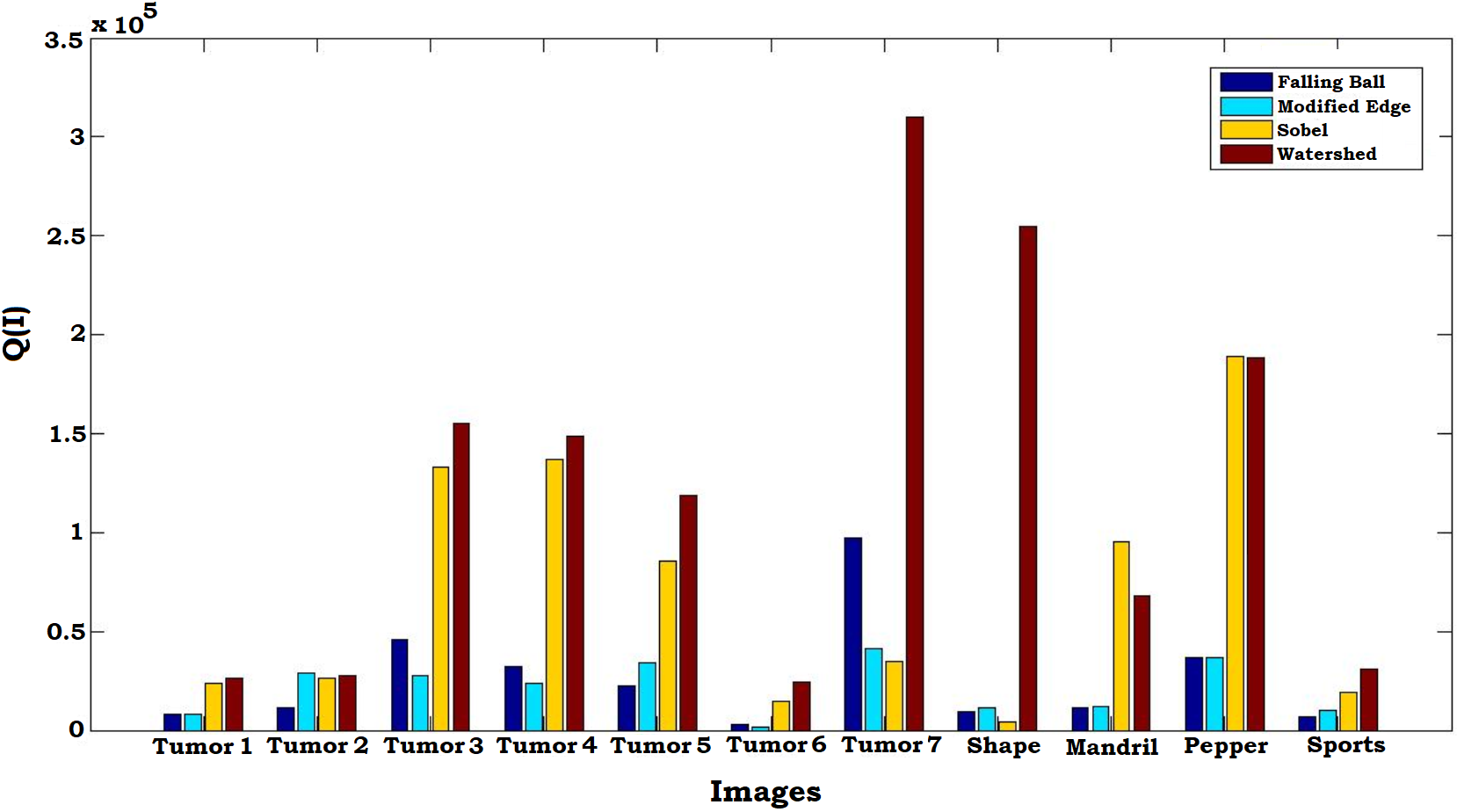}
		\caption{Performance comparison in terms of Q(I) parameter} \label{fig:qi_comparison}
	\end{center}
\end{figure*}
\begin{figure*}[htb]
	\begin{center}
		\includegraphics[width=0.7\textwidth]{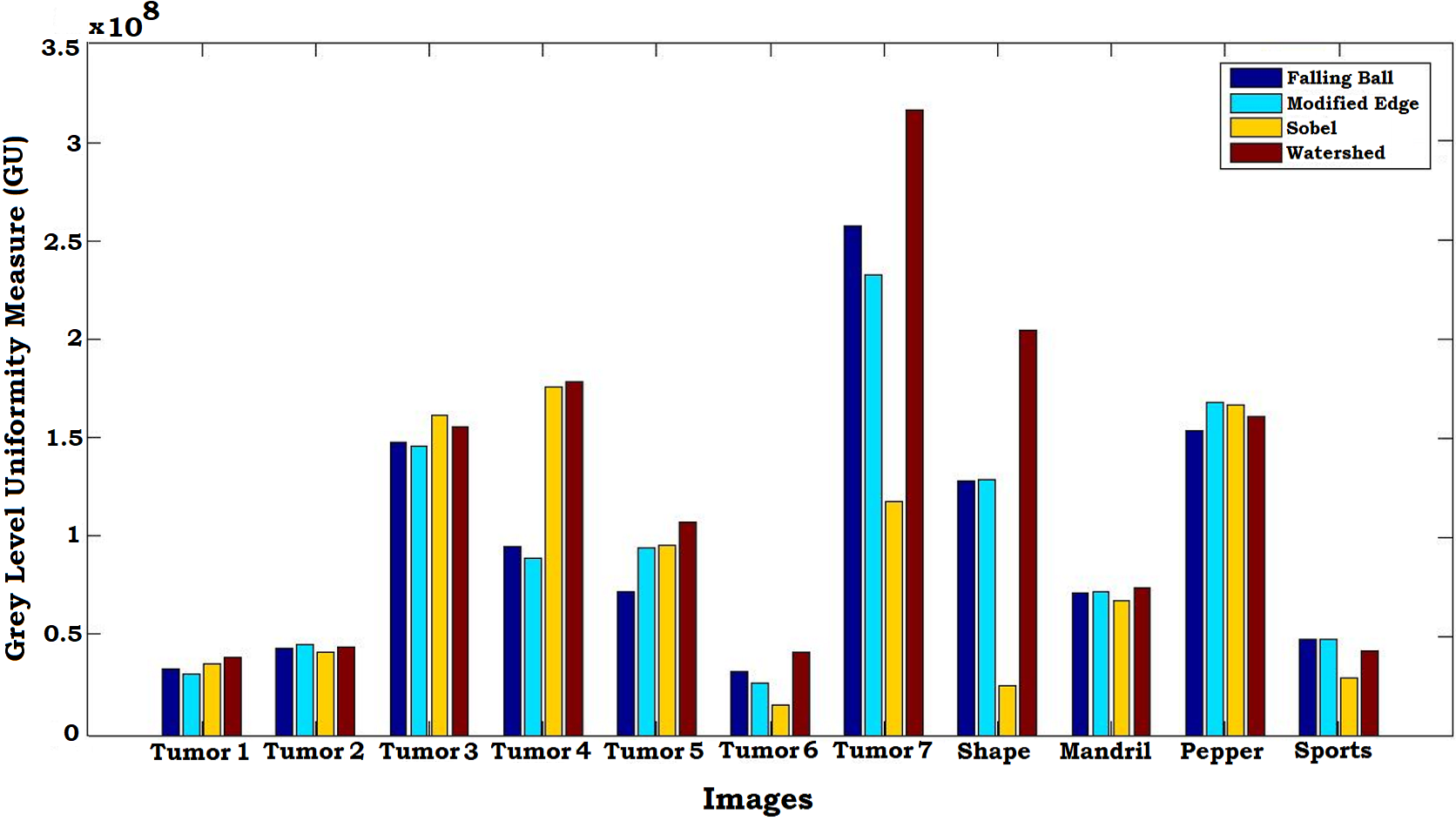}
		\caption{Performance comparison in terms of Grey Level Uniformity measure (GU)} \label{fig:gu_comparison}
	\end{center}
\end{figure*}
\begin{figure*}[tb]
	\begin{center}
		\includegraphics[width=0.7\textwidth]{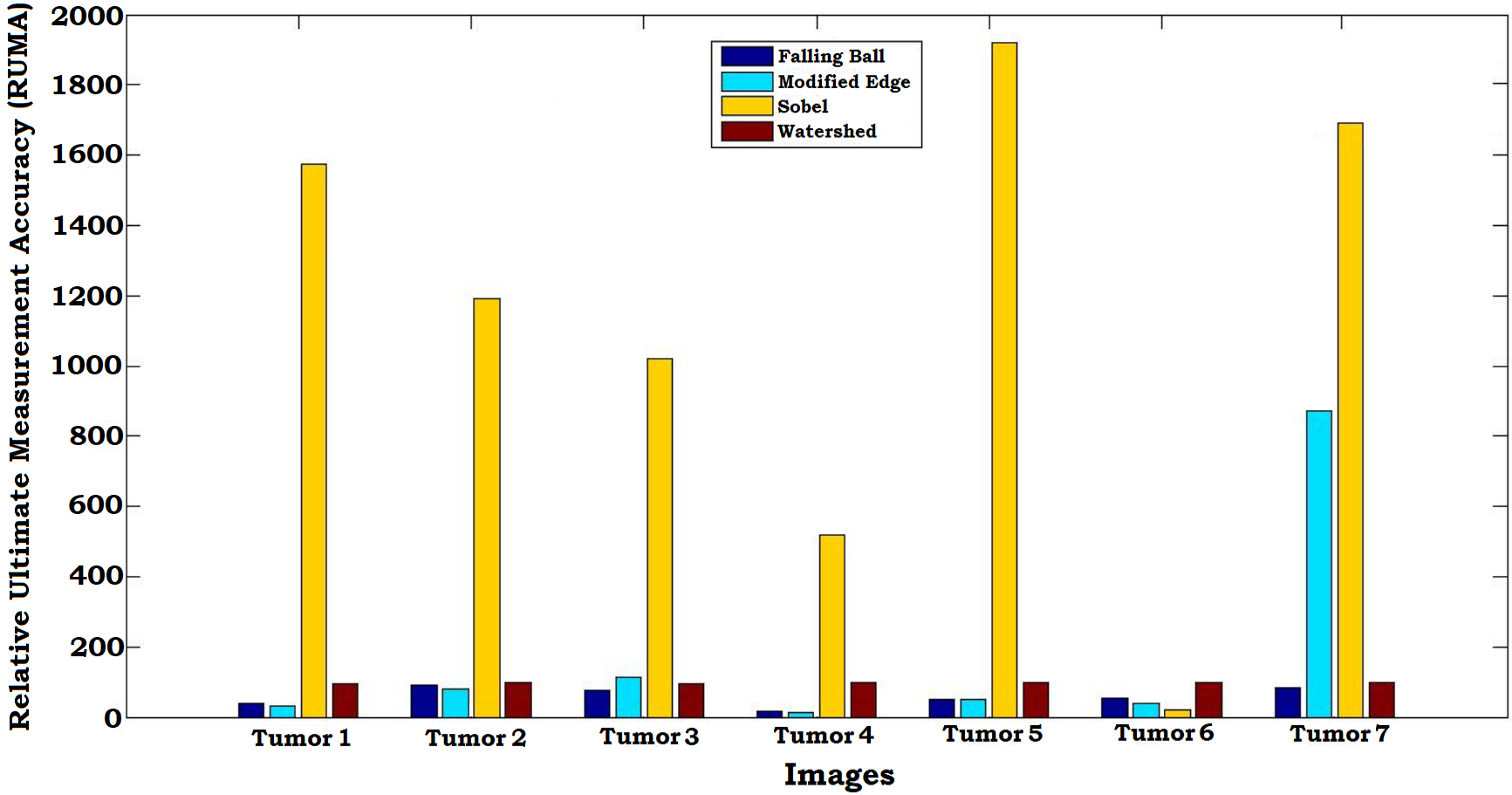}
		\caption{Performance comparison in terms of Relative Ultimate Measurement Accuracy (RUMA)} \label{fig:ruma_comparison}
	\end{center}
\end{figure*}

Furthermore, as the above comparisons only pitted the proposed algorithm against the watershed segmentation method, additional comparisons have been performed to ascertain the performance of the proposed algorithm with respect to other contemporary methods. Fig~\ref{fig:qi_comparison} presents a comparison of the Q-parameter values for four different segmentation methods \textit{viz.} Falling-Ball, Improved-Sobel \cite{aslam2015improved}, standard Sobel, and watershed method. It can be observed that the Falling-Ball algorithm is clearly better than the watershed and Sobel methods, and also outperforms the Improved-Sobel method in 7 out of the 10 images under test. 

Similarly, on the GU criterion, the Falling-Ball algorithm proved to be very potent as can be seen from Fig.~\ref{fig:gu_comparison}. It appears that Tumor-7 and the `shape' images are the outliers in this case, with vastly different GU values for Sobel and watershed algorithms. Lastly, for the RUMA parameter, the comparative analysis is shown in Fig.~\ref{fig:ruma_comparison}. Sobel clearly performs very poorly for all the images in this RUMA test.  The Falling-Ball method turns out to be better than watershed method in all cases, and even outperforms the Improved-Sobel for 4 out of the 7 cases.


For the purpose of subjective quality evaluation, the results of boundaries, region formation and  contours extracted by employing  the watershed and Falling-Ball algorithm on four test images (Tumor-1, Tumor-3, Tumor-4 and Tumor-5) are shown in Fig.~\ref{fig:image1}, \ref{fig:image2}, \ref{fig:image3} and \ref{fig:image4} respectively. Due to scarcity of space, it is not possible to show the results of all images. In Fig.~\ref{fig:image1} through Fig.~\ref{fig:image4}, sub-figure (a) shows the original image, (b) depicts the watershed lines, (c) shows the boundaries detected by proposed algorithm, (d) depicts regions formed by watershed segmentation, (e) shows the regions formed by proposed algorithm, (f) depicts the extracted tumor with watershed method, and lastly (g) shows the extracted tumor with the proposed algorithm.

It is a documented benefit of watershed segmentation that it always gives closed contour and also yields thin edge lines. Although the proposed algorithm requires post-processing with closed contour algorithm, it is more suitable than watershed for segmentation of complex images (like brain tumors). Observing Fig.~\ref{fig:image1}, it can be seen that the regions obtained with watershed segmentation have too many closed contours (Fig.~\ref{fig:image1b}), resulting in false tumor identification (Fig.~\ref{fig:image1f}). On the other hand, the proposed Falling-Ball method generates closed contours (Fig.~\ref{fig:image1e}), which are separated from each other by fine boundaries and different colors. Thus, the  tumor extracted (Fig.~\ref{fig:image1g}) using the proposed method is very identical to the original tumor.

Similar effects are observed in Fig.~\ref{fig:image2}, in which the regions generated by watershed segmentation (Fig.~\ref{fig:image2d}) are over-sensitive like in the previous example. Therefore, the extracted brain tumor  (Fig.~\ref{fig:image2f}) from watershed regions comprises only of a single dot in place of the brain tumor. The regions generated by the proposed method (Fig.~\ref{fig:image2e}) are better than watershed, and therefore it is able to successfully identify the tumor (Fig.~\ref{fig:image2g}), and consequently generate a tumor shape which is more close to the original one. Finally, similar results are obtained in Fig.~\ref{fig:image3} and Fig.~\ref{fig:image4}.

The edges detected by watershed are very sensitive to  contours, and therefore a large number of close contours are inadvertently generated. It is not trivial to detect the location of a tumor correctly from the large number of contours. Contrary to this, although the results of the proposed Falling-Ball algorithm are not always closed contours in intermediate steps,  they are  closed near the brain-tumor. So after applying the closed contour algorithm \cite{aslam2015improved} to the detected edges, tumors can be extracted from the original image.

\section{Conclusions}
\label{sec:conclusion}

In this paper, a novel algorithm was proposed for image segmentation. It was based on fuzzy multi-criteria decision making. This method segments an image into regions using the concept of fuzzy logic with the aid of the movements of a hypothetically falling ball. Regions are constructed from edges using a closed contour algorithm. Although the proposed Falling-Ball algorithm uses some  concepts of watershed algorithm, it is superior than watershed segmentation in terms of extraction of objects of interest. Although watershed gives closed contours, they are over sensitive in nature, thus not very identical to original objects. The Falling-Ball algorithm not only outperformed the watershed method, but also decreased the demerits of a  previously proposed improved-sobel edge detection method \cite{aslam2015improved}. Results were shown for brain tumor MR images. The edges generated by proposed method had less false edges and have closed contours, and the brain tumors extracted  were better than the tumors extracted using watershed segmentation. Both numerical and subjective comparisons demonstrated the superiority of the proposed method over existing counterparts.

As is the case with most research works, the work proposed in this paper is not without its share of limitations. The classifier used  in this work is a very simplified one. A better classifier is expected to yield better results of the Falling-Ball algorithm. Modifications in the fuzzy membership function used in the identification of catchment basins may be explored to improve results. Also, the calculation of thresholding constant, which was taken as 0.11 in this work, also requires  careful attention. A bigger dataset of brain tumor images may help in the refinement of the value of this constant. Also, for images other than those of brain tumors, the calculation of the thresholding constant may be made application specific.


\begin{landscape}
\begin{figure*}[!h]
	\begin{subfigure}{.142\textwidth}
  		\centering
  		\includegraphics[width=.75\linewidth]{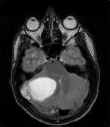}
  		\caption{}
  		\label{fig:image1a}
	\end{subfigure}%
	\hfill
	\begin{subfigure}{.142\textwidth}
		\centering
		\includegraphics[width=.75\linewidth]{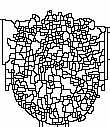}
		\caption{}
		\label{fig:image1b}
	\end{subfigure}%
	\hfill
	\begin{subfigure}{.142\textwidth}
  		\centering
  		\includegraphics[width=.75\linewidth]{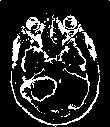}
  		\caption{}
  		\label{fig:image1c}
	\end{subfigure}%
	\hfill
	\begin{subfigure}{.142\textwidth}
  		\centering
  	   \includegraphics[width=.75\linewidth]{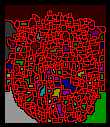}
  		\caption{}
  		\label{fig:image1d}
	\end{subfigure}%
	\hfill
	\begin{subfigure}{.142\textwidth}
		\centering
		\includegraphics[width=.75\linewidth]{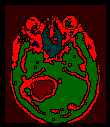}
		\caption{}
		\label{fig:image1e}
	\end{subfigure}%
	\hfill
	\begin{subfigure}{.142\textwidth}
  		\centering
  		\includegraphics[width=.75\linewidth]{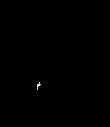}
  		\caption{}
  		\label{fig:image1f}
	\end{subfigure}%
	\hfill
	\begin{subfigure}{.142\textwidth}
  		\centering
  		\includegraphics[width=.75\linewidth]{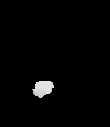}
  		\caption{}
  		\label{fig:image1g}
	\end{subfigure}
\caption{(a) Original Image(Tumor-1), (b) Boundaries formed using Watershed, and (c) Falling-Ball, (d) Regions formed using Watershed, and (e) Falling-Ball, (f) Extracted Tumor from Watershed, and (g) Falling-Ball}
\label{fig:image1}
\end{figure*}
\begin{figure*}[!h]
	\begin{subfigure}{.14\textwidth}
  		\centering
  		\includegraphics[width=.74\linewidth]{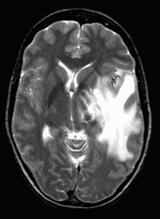}
  		\caption{}
  		\label{fig:image2a}
	\end{subfigure}%
	\hfill
	\begin{subfigure}{.14\textwidth}
		\centering
		\includegraphics[width=.74\linewidth]{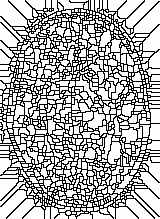}
		\caption{}
		\label{fig:image2b}
	\end{subfigure}%
	\hfill
	\begin{subfigure}{.14\textwidth}
  		\centering
  		\includegraphics[width=.74\linewidth]{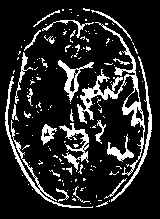}
  		\caption{}
  		\label{fig:image2c}
	\end{subfigure}%
	\hfill
	\begin{subfigure}{.14\textwidth}
  		\centering
  	   \includegraphics[width=.74\linewidth]{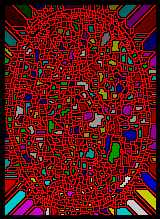}
  		\caption{}
  		\label{fig:image2d}
	\end{subfigure}%
	\hfill
	\begin{subfigure}{.14\textwidth}
		\centering
		\includegraphics[width=.74\linewidth]{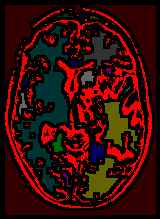}
		\caption{}
		\label{fig:image2e}
	\end{subfigure}%
	\hfill
	\begin{subfigure}{.14\textwidth}
  		\centering
  		\includegraphics[width=.74\linewidth]{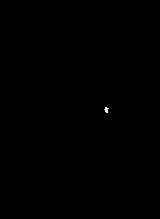}
  		\caption{}
  		\label{fig:image2f}
	\end{subfigure}%
	\hfill
	\begin{subfigure}{.14\textwidth}
  		\centering
  		\includegraphics[width=.74\linewidth]{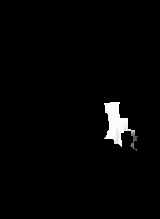}
  		\caption{}
  		\label{fig:image2g}
	\end{subfigure}
\caption{(a) Original Image(Tumor-3), (b) Boundaries formed using Watershed, and (c) Falling-Ball, (d) Regions formed using Watershed, and (e) Falling-Ball, (f) Extracted Tumor from Watershed, and (g) Falling-Ball}
\label{fig:image2}
\end{figure*}
\begin{figure*}[!h]
	\begin{subfigure}{.14\textwidth}
  		\centering
  		\includegraphics[width=.75\linewidth]{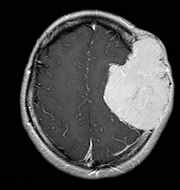}
  		\caption{}
  		\label{fig:image3a}
	\end{subfigure}%
	\hfill
	\begin{subfigure}{.14\textwidth}
		\centering
		\includegraphics[width=.75\linewidth]{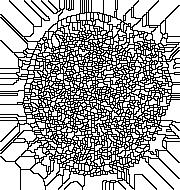}
		\caption{}
		\label{fig:image3b}
	\end{subfigure}%
	\hfill
	\begin{subfigure}{.14\textwidth}
  		\centering
  		\includegraphics[width=.75\linewidth]{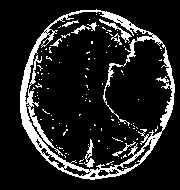}
  		\caption{}
  		\label{fig:image3c}
	\end{subfigure}%
	\hfill
	\begin{subfigure}{.14\textwidth}
  		\centering
  	   \includegraphics[width=.75\linewidth]{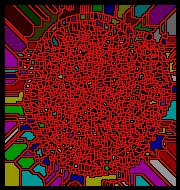}
  		\caption{}
  		\label{fig:image3d}
	\end{subfigure}%
	\hfill
	\begin{subfigure}{.14\textwidth}
		\centering
		\includegraphics[width=.75\linewidth]{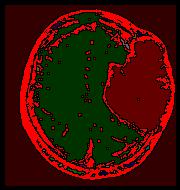}
		\caption{}
		\label{fig:image3e}
	\end{subfigure}%
	\hfill
	\begin{subfigure}{.14\textwidth}
  		\centering
  		\includegraphics[width=.75\linewidth]{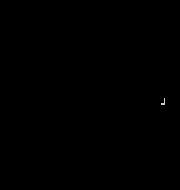}
  		\caption{}
  		\label{fig:image3f}
	\end{subfigure}%
	\hfill
	\begin{subfigure}{.14\textwidth}
  		\centering
  		\includegraphics[width=.75\linewidth]{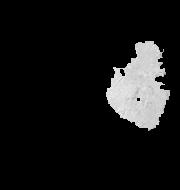}
  		\caption{}
  		\label{fig:image3g}
	\end{subfigure}
\caption{(a) Original Image(Tumor-4), (b) Boundaries formed using Watershed, and (c) Falling-Ball, (d) Regions formed using Watershed, and (e) Falling-Ball, (f) Extracted Tumor from Watershed, and (g) Falling-Ball}
\label{fig:image3}
\end{figure*}
\begin{figure*}
	\begin{subfigure}{.14\textwidth}
		\centering
		\includegraphics[width=.75\linewidth]{images/brain5.jpg}
		\caption{}
		\label{fig:image4a}
	\end{subfigure}%
	\hfill
	\begin{subfigure}{.14\textwidth}
		\centering
		\includegraphics[width=.75\linewidth]{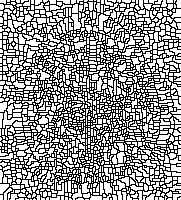}
		\caption{}
		\label{fig:image4b}
	\end{subfigure}%
	\hfill
	\begin{subfigure}{.14\textwidth}
		\centering
		\includegraphics[width=.75\linewidth]{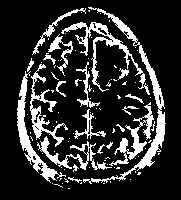}
		\caption{}
		\label{fig:image4c}
	\end{subfigure}%
	\hfill
	\begin{subfigure}{.14\textwidth}
		\centering
		\includegraphics[width=.75\linewidth]{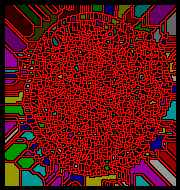}
		\caption{}
		\label{fig:image4d}
	\end{subfigure}%
	\hfill
	\begin{subfigure}{.14\textwidth}
		\centering
		\includegraphics[width=.75\linewidth]{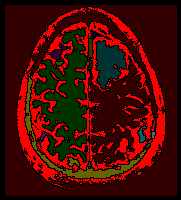}
		\caption{}
		\label{fig:image4e}
	\end{subfigure}%
	\hfill
	\begin{subfigure}{.14\textwidth}
		\centering
		\includegraphics[width=.75\linewidth]{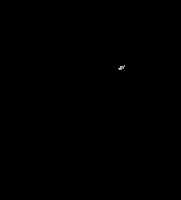}
		\caption{}
		\label{fig:image4f}
	\end{subfigure}%
	\hfill
	\begin{subfigure}{.14\textwidth}
		\centering
		\includegraphics[width=.75\linewidth]{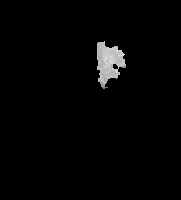}
		\caption{}
		\label{fig:image4g}
	\end{subfigure}
	\caption{(a) Original Image(Tumor-5), (b) Boundaries formed using Watershed, and (c) Falling-Ball, (d) Regions formed using Watershed, and (e) Falling-Ball, (f) Extracted Tumor from Watershed, and (g) Falling-Ball}
	\label{fig:image4}
\end{figure*}
\end{landscape}

\balance
\bibliographystyle{unsrt}
\bibliography{references}

\end{document}